\documentclass[lettersize,journal]{IEEEtran}
\usepackage{amsmath,amsfonts}
\usepackage{algorithmic}
\usepackage{algorithm}
\usepackage{array}
\usepackage[caption=false,font=normalsize,labelfont=sf,textfont=sf]{subfig}
\usepackage{textcomp}
\usepackage{stfloats}
\usepackage{url}
\usepackage{verbatim}
\usepackage{graphicx}
\usepackage{cite}
\usepackage{multirow}
\usepackage{color}
\usepackage{scalerel}
\usepackage{enumerate}
\usepackage{graphicx}
\usepackage{amssymb}
\usepackage{booktabs}
\usepackage{layouts}
\usepackage{subfig}
\usepackage{hyperref}
\usepackage{makecell}

\usepackage{amssymb}
\newcommand\Tau{\mathcal{T}}
\def\etal{\emph{et al}. }

\usepackage[capitalize]{cleveref}
\crefname{section}{Sec.}{Secs.}
\Crefname{section}{Section}{Sections}
\Crefname{table}{Table}{Tables}
\crefname{table}{Tab.}{Tabs.}

\begin{document}

\title{CityTrack: Improving City-Scale Multi-Camera Multi-Target Tracking by Location-Aware Tracking and Box-Grained Matching}

\author{Jincheng Lu\quad
Xipeng Yang\quad
Jin Ye\quad
Yifu Zhang\quad
Zhikang Zou\quad
Wei Zhang\quad
Xiao Tan\quad
\thanks{Jincheng Lu, Xipeng Yang, Yifu Zhang, Wei Zhang, and Xiao Tan are with the Department of Computer Vision, Baidu Inc., Beijing, China. (E-mail: \{lujincheng01, yangxipeng01, zhangyifu, zhangwei99, tanxiao01\}@baidu.com). Jin Ye is with Shanghai AI Laboratory, Shanghai, China. (Email: yejin16@mails.ucas.ac.cn). The first two authors contributed equally to this work.}}



\maketitle

\begin{abstract}
Multi-Camera Multi-Target Tracking (MCMT) is a computer vision technique that involves tracking multiple targets simultaneously across multiple cameras. MCMT in urban traffic visual analysis faces great challenges due to the complex and dynamic nature of urban traffic scenes, where multiple cameras with different views and perspectives are often used to cover a large city-scale area. Targets in urban traffic scenes often undergo occlusion, illumination changes, and perspective changes, making it difficult to associate targets across different cameras accurately. To overcome these challenges, we propose a novel systematic MCMT framework, called CityTrack. Specifically, we present a Location-Aware SCMT tracker which integrates various advanced techniques to improve its effectiveness in the MCMT task and propose a novel Box-Grained Matching~(BGM) method for the ICA module to solve the aforementioned problems. We evaluated our approach on the public test set of the CityFlowV2 dataset and achieved an IDF1 of 84.91\%, ranking \textbf{1}st in the 2022 AI CITY CHALLENGE. Our experimental results demonstrate the effectiveness of our approach in overcoming the challenges posed by urban traffic scenes. 
\end{abstract}


\section{Introduction}
\label{sec:intro}
\begin{figure}[t]
  \centering
  \includegraphics[width=0.9\linewidth]{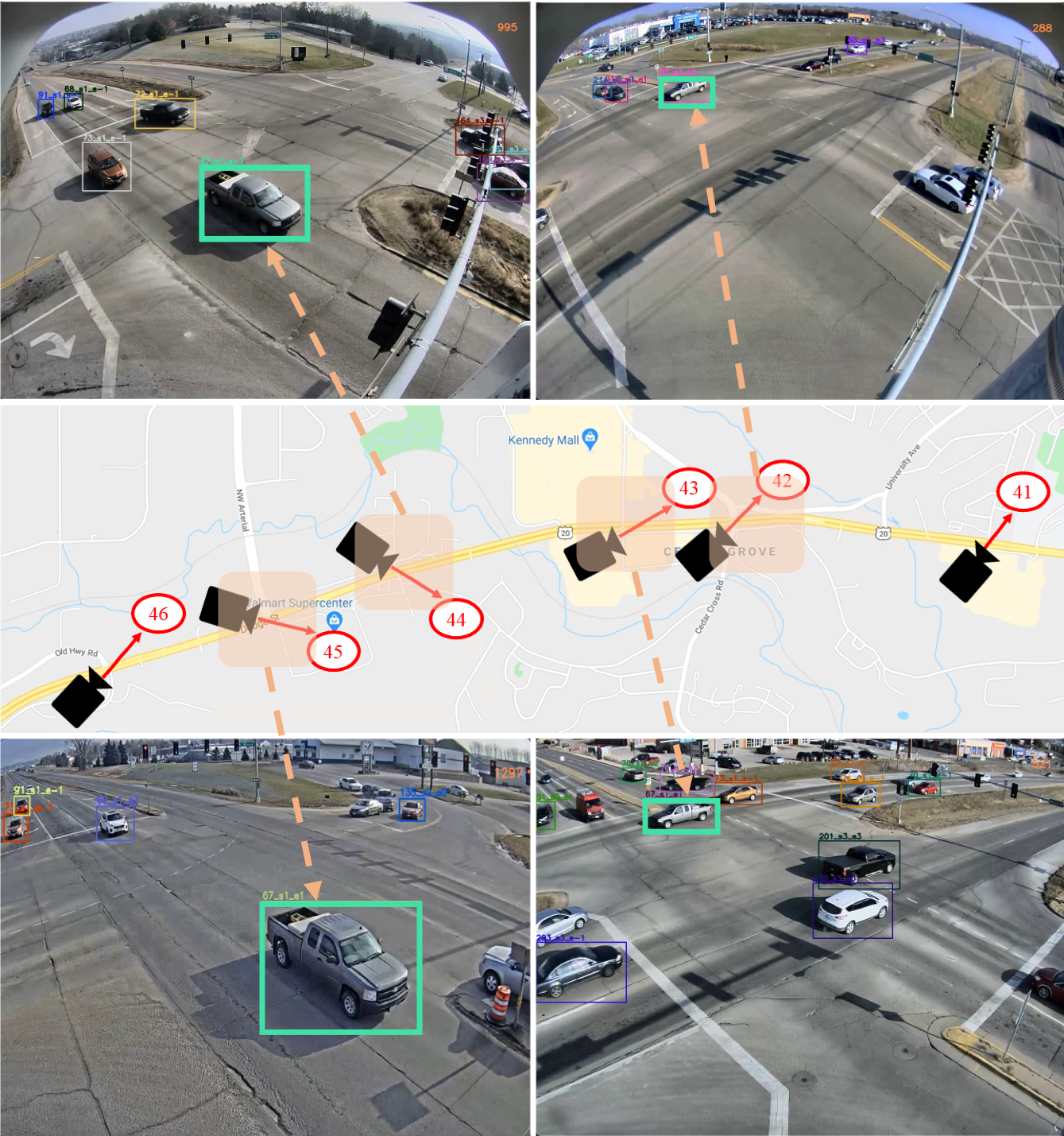}
  \caption{Illustration of Multi-Camera Multi-Target Tracking~(MCMT) task. Our proposed system will match the vehicles with the same identity that appears in multiple cameras.}
  \label{fig:camera-loc}
\end{figure}

With the rapid development of intelligent transportation systems, the demand for Multi-Camera Multi-Target Tracking (MCMT) has attracted extensive attention in recent years. The purpose of the MCMT of vehicles is to track various vehicles across multiple cameras as shown in Figure \ref{fig:camera-loc}, which helps to analyze the traffic flow and travel time along entire corridors. The system designed to tackle the MCMT task typically consists of vehicle detection, Re-Identification~(ReID) Feature Extraction, Single-Camera Multi-Target Tracking (SCMT), and Inter-Camera Association (ICA). The general pipeline can be summarized as follows: First, the vehicle detection module outputs vehicle coordinates and categories in frames and the ReID module extracts the appearance features of vehicles. Then, based on the vehicle location and appearance features, the SCMT module generates trajectories for every single camera. At last, the ICA module matches these candidate trajectories across different cameras to associate vehicles with the same identities.

In the last few years, there are an increasing number of research efforts dedicated to solving the MCMT task~\cite{he2019multi,hou2019locality,hsu2019multi,lee2015combined,Qian_2020_CVPR_Workshops,vspanhel2019vehicle,tan2019multi,tang2019cityflow,tang2018single}. Hsu~\etal~\cite{hsu2019multi} firstly propose many intricate traveling strategies to distinguish tracklets with several pre-defined zones for every camera, and then apply the greedy algorithm to generate the inter-camera matching result. Qian~\etal~\cite{Qian_2020_CVPR_Workshops} propose a rule-based algorithm, which adopts the geometry information to update SCMT tracking results repeatedly. However, the above methods do not directly focus on the MCMT tracking task in intelligent transportation scenarios. Although these MCMT methods achieve competitive tracking results, they all rely on tracklet-grained feature matching. The variance of vehicles' appearance features through multiple cameras is large. It is hard to discriminate the identities when using the mean feature or single frame feature to represent the tracklet. By contrast, we propose a box-grained matching method that takes all the boxes into account when calculating the cost for trajectories association. In addition, in order to avoid error matching of similar vehicles, we use a stricter k-reciprocal nearest neighbor rule for accurate matching. Another remaining challenge is that the common SCMT trackers often generate broken tracking results under heavily occluded traffic scenes, which causes matching ambiguity for the ICA module. To solve this problem, we propose a location-aware SCMT tracker, using three strategies to improve the tracking performance for different locations in traffic scenes, named Stationary Sensitive Association~(SSA), Trajectory Re-Link~(TRL), and Bidirectional Tracking~(BT).

\begin{figure*}[t]
\begin{center}
    \includegraphics[width=0.98\linewidth]{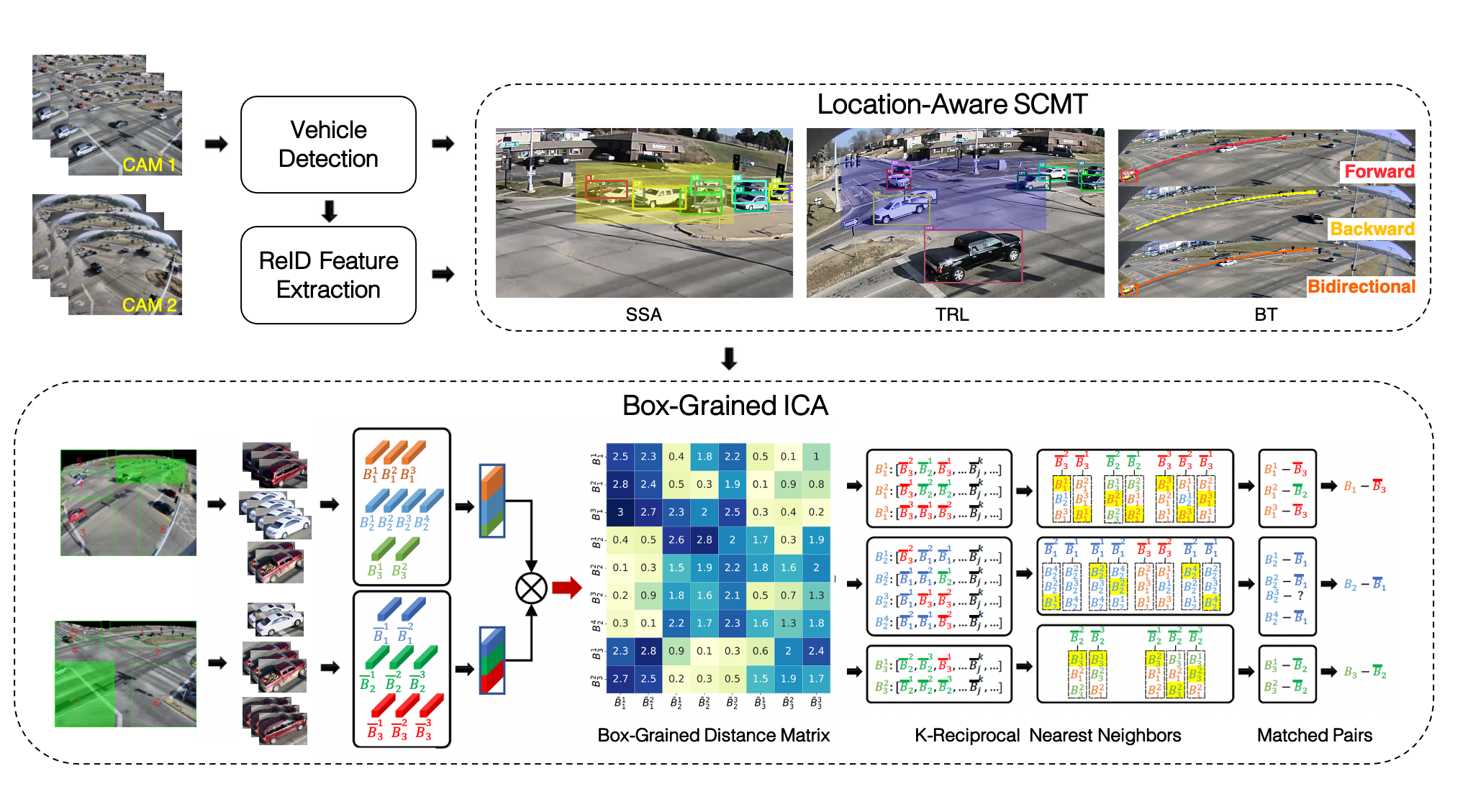}
\end{center}
\caption{The pipeline of our MCMT tracking system. The detected vehicle boxes and their features are fed into the SCMT module, which generates all trajectories for every single camera. Our proposed Location-aware SCMT upgrades the baseline tracker from three aspects. Stationary Sensitive Association~(SSA) improves the tracking of vehicle stopping areas (viewed in yellow), Trajectory Re-Link~(TRL) reduces id switches in the middle of the scenarios (viewed in blue), and Bidirectional Tracking~(BT) completes the trajectories at the beginning and end. Finally, our proposed ICA module matches trajectories across all cameras. Filtered by the pre-defined zones, we obtain exiting vehicles and entering vehicles of two successive cameras. The cosine distances between features of all exiting and entering boxes are calculated to form the box-grained distance matrix. We do trajectories association with the principle of k-reciprocal nearest neighbors to find the final matched pairs.}
\label{fig:framework}
\end{figure*}

In this paper, we propose an entire MCMT tracking system to alleviate tracking challenges of similar appearance and severe occlusions. The pipeline of our proposed MCMT tracking system is shown in Figure \ref{fig:framework}. Given a set of videos under different cameras, our system first detects all vehicles via a detector and extracts the appearance features. We propose a Location-Aware SCMT module. The baseline tracker is equipped with SSA, TRL, and BT strategies to generate accurate and complete trajectories for different locations in traffic scenes. Finally, the Box-Grained ICA module associates all candidate trajectories between two successive cameras. The cosine distances between features of every exiting and entering vehicle box are calculated to form the box-grained distance matrix. We do trajectory association with the principle of k-reciprocal nearest neighbors to find the matched pairs. The proposed system achieves IDF1 84.91\% on the CityFlowV2 dataset, which outperforms
the state-of-the-art MCMT methods. The contributions of our work are summarized as follows:
\begin{enumerate}[{1)}]
\item We present a Location-Aware SCMT tracker that improves the baseline tracker with advanced techniques designed for different locations in traffic scenarios.
\item We propose a novel Box-Grained Matching method for the ICA module that can better discriminate vehicles with similar appearances.
\item By combining the Location-Aware SCMT module and the Box-Grained ICA module, our MCMT system set a new state-of-the-art record on the CityFlowV2 benchmark.
\end{enumerate}


The rest of the paper is organized as follows: an overview of related work is described in Section~\ref{sec:rel}. Section~\ref{sec:method} introduces our proposed system in detail. In Section~\ref{sec:exp}, we demonstrate sufficient experiments of our method and ablation studies. Finally, we present the conclusion in Section~\ref{sec:con}.

\section{Related work}
\label{sec:rel}

\subsection{Vehicle Detection}
Object detection is one of the most popular tasks in the field of computer vision. It locates the existence of objects in an image by predicting the bounding boxes and categories. The vehicle detection task is a special object detection branch. Vehicles tend to have similar appearances but different sizes due to changes in camera views. Due to the rapid development of convolution networks, CNN-based detectors have achieved tremendous progress in the past few years, such as SSD \cite{liu2016ssd}, Yolo \cite{redmon2017yolo9000}, Faster-RCNN \cite{ren2015faster}, Cascade-RCNN \cite{cai2018cascade}. Multi-layer feature fusion networks such as \cite{lin2017feature,liu2018path} play an important role to detect objects of different sizes in the vehicle detection problem. Recently, inspired by the success of transformer \cite{vaswani2017attention} in natural language processing, the newly proposed object detectors, such as DETR \cite{carion2020end}, and SwinTransformer \cite{liu2021swin}, introduced vision transformer-based detectors that achieve competitive performances on object detection benchmarks by treating an image as a sequence of patches. The transformer backbones can obtain long-distance pixel relations better. We adopt the Cascade-RCNN with SwinTransformer backbone as our detector as it handles different sizes of objects and captures global information in crowd scenarios under the vehicle detection setting.

\subsection{ReID Feature Extraction}
As one of the most important components in the MCMT task, ReID Feature Extraction aims to retrieve the same object captured by different cameras \cite{zheng2020vehiclenet}. CNN-based ReID methods have received extensive attention and shown strong feature representation ability. In these methods, several loss functions 
and data generation methods are proposed to learn discriminative feature representation.

There are three commonly used loss functions for ReID, including identity loss, verification loss, and triplet loss \cite{ye2021deep}. By using identity loss such as cross-entropy loss \cite{zhang2018generalized}, the training process of ReID is treated as an image classification problem. Verification loss such as contrastive loss optimizes the pairwise relationship \cite{varior2016siamese} by treating the training process as an image-matching problem. Triplet loss treats the ReID training process as a retrieval ranking problem \cite{hermans2017defense}, aiming to make the distance between positive pairs smaller than negative pairs. We ensemble several ReID models trained with cross-entropy loss and triplet loss to extract more discriminative appearance features. The appearance features are utilized to distinguish different vehicles in the tracking process. 



\subsection{Single-Camera Multi-Target Tracking}
Multi-Target Tracking (or Multi-Object Tracking) plays a crucial role in areas like video understanding, traffic control, and autonomous driving. The goal of Single-Camera Multi-Target Tracking is to associate multiple objects, maintain their identities, and yield their trajectories in an input video sequence. Modern SCMT trackers can be classified as tracking-by-detection methods and joint-detection-tracking methods.

Tracking-by-detection methods \cite{ref_sort, ref_deepsort, ref_poi, ref_qdtrack, zhang2021bytetrack, du2022strongsort, ref_ocsort, ref_botsort, zhang2022robust} obtain detection boxes first and then associate them based on appearance and motion clues. With the improvement of object detection techniques \cite{ren2015faster,he2017mask,redmon2017yolo9000,ref_yolox,liu2021swin}, tracking-by-detection methods have dominated the SCMT task for years. SORT \cite{ref_sort} adopts the Kalman filter algorithm \cite{kalman1960new} for motion prediction and matches the detection boxes to the tracklets by Hungarian Algorithm \cite{ref_hungarian}. DeepSORT \cite{ref_deepsort} introduces deep visual features into data association and proposes a cascaded matching strategy that matches the detection boxes to the tracklets in order based on their lost time for a more reliable association. 

Recently, several joint-detection-tracking methods incorporate appearance embedding or motion prediction into detection frameworks \cite{ref_jde,ref_fairmot,ref_centertrack,ref_cstrack,ref_siammot}. The joint trackers achieve comparable performance with low computational costs. However, joint trackers are facing the problem that the competition between different components lowers the upper bound of tracking performance. 

The success of the latest SORT-like frameworks \cite{zhang2021bytetrack, du2022strongsort, ref_ocsort, ref_botsort} indicates that the tracking-by-detection paradigm is still the optimal solution in terms of tracking accuracy. In order not to miss potential targets with low detection confidence (e.g. occluded vehicles and small vehicles), ByteTrack \cite{zhang2021bytetrack} sets a high score threshold and a low score threshold to filter detection results and perform matching in stages. New tracklets are initiated only for unmatched high-score detections. StrongSORT \cite{du2022strongsort} upgrade DeepSORT from appearance feature updating and motion prediction. The appearance features of the tracklets are updated in an exponential moving average (EMA) manner. For the motion prediction, StrongSORT uses the NSA Kalman filter \cite{du2021giaotracker} to incorporate the confidence of detection into covariance calculation. The final similarity distance is a weighted sum of appearance feature cosine distance and Mahalanobis distance. We follow the tracking-by-detection paradigm and extend \cite{zhang2021bytetrack} and \cite{du2022strongsort} to the vehicle tracking setting. We treat the still and moving vehicles differently to address the nonlinear motion problem in intelligent transportation scenarios.

Offline trackers adopt global information to refine the tracking results further. TPM \cite{ref_tpm} presents a tracklet-plane matching process to push easily confusable tracklets into different tracklet-planes. TNT \cite{ref_tnt}, GIAOTracker \cite{du2021giaotracker} and StrongSORT \cite{du2022strongsort} employ CNN models to link tracklets based on the motion and appearance information. It often takes a cost of more computation time but achieves better performance. We propose a Trajectory Re-Link strategy to merge broken trajectories to address the vehicle occlusion problem.

\subsection{Inter-Camera Association}
After obtaining all results from the above three modules, the inter-camera association can be treated as trajectories matching or tracklets retrieval problem. Many previous works attempt to tackle this problem from different aspects. Chen~\etal~\cite{chen2014novel,chen2016equalized} establish a global graph for multiple tracklets in different cameras and optimize for an MCMT solution. 
Recently, many works~\cite{hsu2019multi,hsu2021multi,liu2021city,tang2018single,ye2021robust} find that traffic rules and spatial-temporal constraints can be regarded as the prior knowledge to filter out the tracklets candidates, which reduces the searching space significantly. After the preprocessing, Hsu~\etal~\cite{hsu2019multi} uses the greedy algorithm to search the valid tracklet pairs. Ye~\etal~\cite{ye2021robust} adopts the Hungarian matching algorithm to find the global optimization results with the distance matrix of all tracklets candidates between two successive cameras. Liu~\etal~\cite{liu2021city} introduces hierarchical clustering to gather potential trajectory pairs within two cameras. Unlike existing works that adopt tracklet-grained features, we construct the distance matrix with box-grained features to handle similar appearances problem. We also obtain more accurate tracklet pairs in a k-reciprocal way.

\section{Methodology}
\label{sec:method}

\subsection{Problem Setup}
The purpose of the MCMT system is to track the vehicles in multi-cameras and generate the entire trajectories. For each input video frame, the detection module predicts the bounding boxes. The ReID module extracts the features for each detection. The SCMT module generates trajectories $\Tau$ by associating detection boxes to the tracklets from previous frames. Each trajectory is composed of detected bounding boxes and their features, denoted as $\tau_{i}=[B_{1}, B_{2}, ..., B_{l_{i}}]$, where $l_{i}$ represents the length of the i-th trajectory, and $B=[x, y, w, h, f]$ is a bounding box composed of its coordinates and ReID feature. With the exiting trajectories (which end in the exit zone) $\Tau_{out}=[\tau_{1}, \tau_{2}, ..., \tau_{n}]$ and the entering trajectories (which start in the entrance zone) $\Tau_{in}=[\bar{\tau}_{1}, \bar{\tau}_{2}, ..., \bar{\tau}_{m}]$ of every two successive cameras as input, the ICA module solves the matching problem and assigns global identities for each trajectory.

\subsection{Overview of CityTrack}
This section presents the details of the proposed CityTrack MCMT tracker. Figure \ref{fig:framework} illustrates the system's framework, which consists of four modules: Detection, ReID Feature Extraction, SCMT, and ICA. The detection module is responsible for detecting all vehicle objects in the scene, and the ReID module extracts the corresponding appearance features for each object. In our system, we use an off-the-shelf object detector and ReID feature extraction network to perform these tasks. The SCMT module receives the detected boxes and their features as input and generates individual trajectories for each object within a single camera's view. Finally, the ICA module matches trajectory candidates across all cameras to produce the final MCMT result.

\subsection{Single-Camera Multi-Target Tracking}

Provided with high-quality detection results and ReID features, our Single-Camera Multi-Target Tracking~(SCMT) module focuses on associating targets throughout the video frames following the tracking-by-detection paradigm. We integrate the data association method BYTE\cite{zhang2021bytetrack} into the StrongSORT\cite{du2022strongsort} as our baseline tracker. In the MCMT task, broken trajectories in the SCMT stage are especially detrimental, because the extra candidate trajectories can cause incomplete results and false matches in the ICA stage. To address the challenges posed by nonlinear motion, severe occlusion, and imperfect detection in real-world traffic, we have designed a Location-Aware SCMT tracker to improve tracking performance in crowded scenarios. Specifically, we present three strategies: Stationary Sensitive Association~(SSA) improves the tracking in areas where vehicles are stopped, Trajectory Re-Link~(TRL) reduces identity switches in the middle of the scenarios, and Bidirectional Tracking~(BT) completes the trajectories at the beginning and end. While we do not introduce any novel algorithms in this section, our contributions include incorporating these advanced techniques to enhance the effectiveness of the baseline tracker in the MCMT task.

\begin{figure}
	\centering
	\subfloat[]{
		\includegraphics[width=0.23\textwidth]{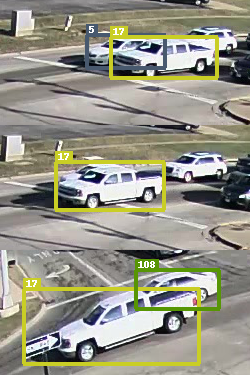} \label{fig:offline_merge1}}
	\subfloat[]{
		\includegraphics[width=0.23\textwidth]{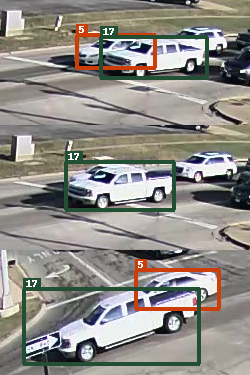} \label{fig:offline_merge2}}
		
\caption{Comparison of tracking results before and after using the Trajectory Re-Link strategy. (a) is the result that the ID of the occluded white car switched from 5 to 108. (b) is the result of using Trajectory Re-Link.}
\label{fig:offline_merge}
\end{figure}

\textbf{Stationary Sensitive Association} Vehicles will stop and start at the intersections. We address the nonlinear motion problem by refining the detection and adjusting motion constraints. If a tracklet and tracklets around it all remain stationary, we keep the detection with the highest score as its location to handle the loss of detection caused by occlusion. To further improve the robustness of nonlinear motion, we borrow the smoothed Mahalanobis distance from \cite{liu2020robust} to avoid sudden change during the non-linear motion stage.

\textbf{Trajectory Re-Link} Object detection can be difficult under severe occlusion, as shown in Figure \ref{fig:offline_merge1}, where the track ID of the white car switches after it becomes occluded. This is because the sharp change in velocity when the car starts causes the Kalman filter to be unable to accurately predict motion states. To address this problem, we first filter out trajectories that end or start in the middle of the scene. Then, we use a greedy algorithm to merge broken trajectories based on the cosine distance of their ReID features. As shown in Figure \ref{fig:offline_merge2}, the Trajectory Re-Link strategy is able to maintain the same track identity for the white car even after it has been occluded.

\textbf{Bidirectional Tracking} Although we have adopted various techniques to reduce ID switches, the trajectories can still be incomplete due to imperfect detection results. The targets away from the camera tend to be ignored at the beginning frames because the detection confidences of small targets are too low to initiate tracklets. Tracking backward can solve the problem by initiating tracklets in the area close to the camera. Similar to \cite{stadler2021improving}, by running our tracker on the video frames one time in the forward direction, and one time in the backward direction, we merge the tracklets to generate complete trajectories.

\begin{figure}
	\centering
	\includegraphics[width=0.45\textwidth]{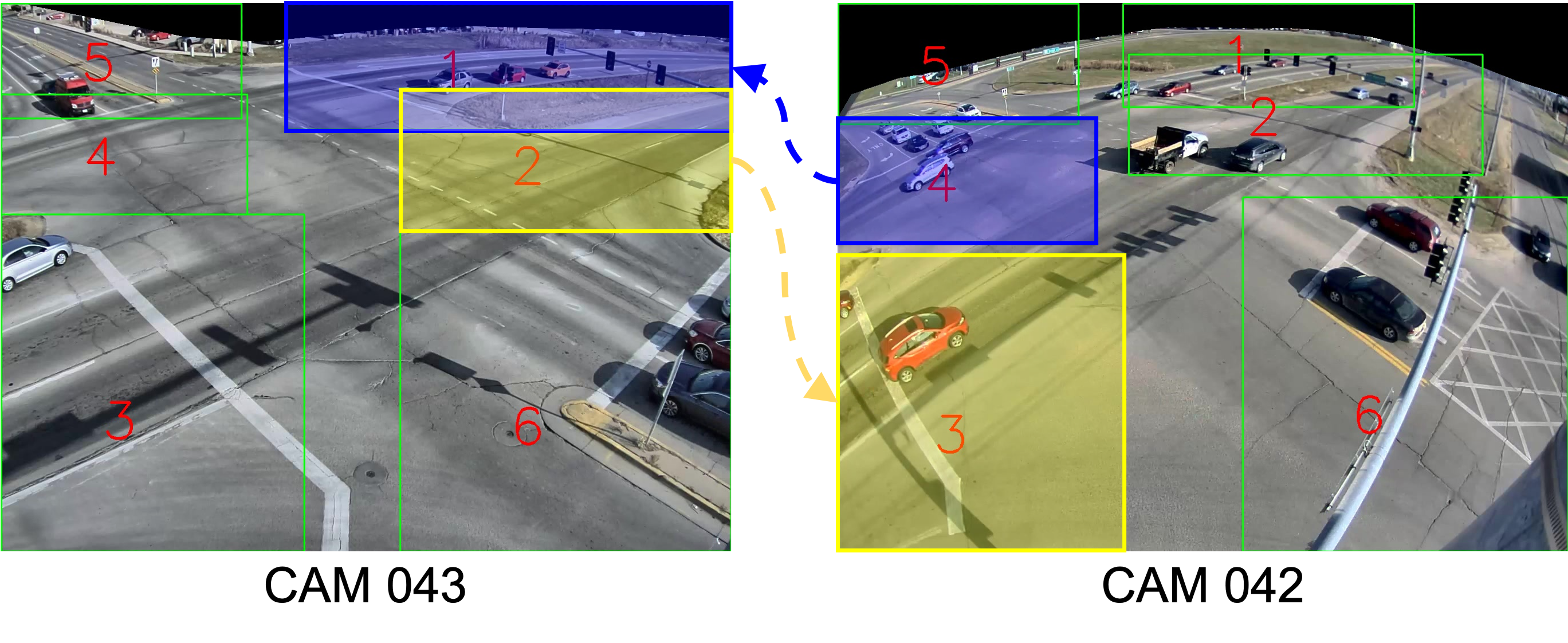}

\caption{Examples of predefined zones to describe trajectories. The valid exiting-entering zone pairs are viewed in the same color.}
\label{fig:zones}
\end{figure}

\subsection{Inter-Camera Association}
The Inter-Camera Association (ICA) module is the final and most critical component of the MCMT system. It uses the trajectories generated by the previous modules to associate all trajectories with the same identities based on appearance features and spatial-temporal information. The ICA module matches the trajectories of two consecutive cameras based on the entrance and exit of the road. However, there are several challenges that the ICA module must overcome, such as the presence of multiple vehicles with similar appearances in the pool of trajectory candidates, which can lead to errors in the matching process. In addition, differences in camera locations can affect factors such as illumination and perspective, making the matching task more difficult. To address these challenges, we propose a novel Box-Grained Matching (BGM) module that identifies the same identities at the box level, rather than the tracklet-grained matching strategy used in previous methods. In the following section, we will first introduce the tracklet-grained baseline method and then discuss our proposed BGM module in more detail.

\subsubsection{Baseline}
\ 
\newline
\indent The tracklet-grained baseline method first constructs a pool of trajectories to match based on spatial-temporal information, then calculates the tracklet-grained distance matrix and performs trajectory association to obtain the MCMT result.

\textbf{Trajectories Pool Construction.} 
Provided with trajectories in each camera, \cite{ye2021robust} use the spatial-temporal information to construct a clean candidate trajectories pool for ICA. Spatial and temporal constraints are used to filter out invalid trajectories. The first step is to filter the trajectories based on the predefined zones and the road topology, as shown in Figure \ref{fig:zones}. Only the trajectories passing through valid zone pairs~(e.g. zone 4 of C042 as the exiting zone and zone 1 of C043 as the entering zone) are considered in the matching process. The second step is to filter out trajectories that cannot appear in adjacent cameras based on temporal information. Because adjacent cameras are spaced apart, not all vehicles that appear in an upstream camera can reach a downstream camera. We obtain the valid exiting trajectories $\Tau_{out}$ and entering trajectories $\Tau_{in}$ by the following formulas:
\begin{small}
\begin{align}
\label{equ:PoolOut}
    & \Tau_{out} = [\Tau_{out}(t_{out} < T_{out})\cap \Tau_{out}(z_{out})] \\
\label{equ:PoolIn}
    & \Tau_{in} = [\Tau_{in}(t_{in} > T_{in})\cap \Tau_{in}(z_{in})]
\end{align}
\end{small}
where $t_{out}$ and $t_{in}$ are the timestamp when the tracklet exits and enters the camera, $T_{out}$ and $T_{in}$ are the thresholds of timestamp for exiting and entering cameras, $z_{out}$ and $z_{in}$ are the exiting zone and the entering zone of the cameras. After the Trajectories Pool Construction, the search space is significantly reduced. 

\textbf{Distance Matrix Construction.}
After finishing candidate trajectories selection, it uses the mean of the minimum k cosine distance between trajectories to calculate the similarity distance of the trajectories $\tau_{i}\in{\Tau_{out}}$, $\bar{\tau_{j}}\in{\Tau_{in}}$ as follows:
\begin{equation}
    d\_traj(\tau_i, \bar{\tau}_j) = \frac{1}{k}\sum_{l=1}^{k}min\_{l}(1 - cos(B_i, \bar{B}_j))
    \label{eq:two_feature_dist}
\end{equation}
where we denote $min\_{l}$ as the $l$-th minimum value, $B_i$ and $\bar{B_{j}}$ as the bounding box features of $\tau_{i}$ and $\tau_{j}$. The tracklet-grained similarity distance matrix $D$ is constructed as:
\begin{small} 
\begin{gather}
\label{equ:disMatTracklet}
D = \begin{bmatrix}
    d\_traj(\tau_1, \bar{\tau}_1) & \dots & d\_traj(\tau_1, \bar{\tau}_m) \\
    \vdots & \ddots & \vdots \\
    d\_traj(\tau_n, \bar{\tau}_1) & \dots & d\_traj(\tau_n, \bar{\tau}_m)
\end{bmatrix}_{n \times m}
\end{gather}
\end{small}
$n$ is the size of $\Tau_{out}$ and $m$ is the size of $\Tau_{in}$.

If the traveling time is out of the valid time window, we multiply the distance of the pair by a penalty factor:
\begin{small}
\begin{align}
D_{i,j} = 
&\begin{cases}
    {\alpha}{\times}D_{i,j}, & t_{i,j} \leq T_{low} \\
    D_{i,j}, & T_{low} < t_{i,j} < T_{upp} \\
    {\alpha}{\times}D_{i,j}, & t_{i,j} \geq T_{upp}
\end{cases}
\end{align}
\end{small}
where $T_{low}$ and $T_{upp}$ are the thresholds of the traveling time window, $t_{i,j}$ is the interval time of $\tau_i$ and $\bar{\tau}_j$, and $\alpha$ is the penalty factor for those pairs outside the time window, which equals 2 as in \cite{ye2021robust}.

\textbf{Trajectory Association.}
Finally, the Hungarian algorithm is performed to associate trajectories. After getting matched results, we check the temporal validity of all matched pairs to remove infeasible pairs where the time of the exiting zone is earlier than the time of the entering zone. Then, we merge all pairs with the same global ID to obtain global trajectories.

\subsubsection{Box-Grained Matching}
\ 
\newline
\indent Algorithm \ref{alg:ICA} outlines the pipeline of our proposed Box-Grained ICA module. The ``Trajectories Pool Construction'' step is the same as in the baseline method.

\textbf{Box-Grained Distance Matrix Construction.}
Once $\Tau_{out}$ set and $\Tau_{in}$ set are obtained, previous methods~\cite{liu2021city,ye2021robust} calculate the tracklet-grained distance matrix for the final matching. This way only can get limited performance due to some noisy appearance features within tracklets that may dominate their representations. To solve this problem, we calculate the box-grained distance matrix instead. Take two connected zones into account, before starting to match, we need to calculate the distance between each box for the two zones. $\tau_i=[B_i^1,...,B_i^{l_i}]$ and $\bar{\tau}_j=[\bar{B}_j^1,...,\bar{B}_j^{l_j}]$ are the tracklets of the exiting zone and the entering zone, respectively. $B_i^l$ is the $l$th box feature of tracklet $i$. From this we can get the similarity distance matrix $D$ between the two zones:
\begin{small}
\begin{gather}
\label{equ:disMat}
D = \begin{bmatrix}
    1-cos(B_1^1,\bar{B}_1^1) & \dots & 1-cos(B_1^{1}, \bar{B}_m^{l_m}) \\
    \vdots & \ddots & \vdots \\
    1-cos(B_n^{l_n}, \bar{B}_1^1) & \dots & 1-cos(B_n^{l_n}, \bar{B}_m^{l_m})
\end{bmatrix}_{\sum\limits_{i=1}^{n}{l_i} \times \sum\limits_{j=1}^{m}{l_j} }
\end{gather}
\end{small}
where $B_i^l$ is the $l$th~($l\in[0,l_i]$) box feature of tracklet $i$ of exiting zone, $\bar{B}_j^l$ is the $l$th~($l\in[0,l_j]$) box feature of tracklet $j$ of entering zone. 

The similarity matrix may still need to be optimized due to severe occlusion, illumination, or different view perspective. To improve the reliability of the similarity matrix, we focus on adjusting the weights among boxes in three steps: re-ranking, spatio-temporal prior information, and occlusion size. First, we use the re-ranking method~\cite{zhong2017re} to reconstruct the similarity matrix $D$ of distance matrix $\hat{D}$. This helps to improve the matching performance by considering the overall context of the boxes, rather than just their individual features.

Guided with the temporal prior information, we refine $\hat{D}$ with interval thresholds, 
\begin{large}
\begin{align}
    \hat{D}_{i,j} = &\begin{cases}
    e^{\frac{\alpha_t\times(T_{low} - t_{i,j})}{\beta_t}}\times{\hat{D}_{i,j}}, & t_{i,j} < T_{low} \\
    e^{\frac{\alpha_t\times(t_{i,j} - T_{upp})}{\beta_t}}\times{\hat{D}_{i,j}}, & t_{i,j} > T_{upp} \\
\end{cases}
\end{align}
\end{large}
where $D_{i,j}$ is the distance between $i$ and $j$ in the distance matrix, $\alpha_t$ and $\beta_t$ are the hyperparameters, $t_{low}$ and $t_{upp}$ are the lower threshold and upper threshold of traveling time window, respectively.

In the last step, we refine the distance matrix with occlusion rate to generate a convincing distance matrix $D$ for final matching,
\begin{align}
    \label{equ:occRefine}
    \hat{D}_{i, j} = &\begin{cases}
    e^{\alpha_o\times(1 + r_o)}\times{\hat{D}_{i,j}}, & r_o > r_{thre} \\
    \hat{D}_{i, j}, &otherwise \\
\end{cases}
\end{align}
$\alpha_o$ is the hyperparameter, $r_o$ and $r_{thre}$ are the occlusion rate for box $i$ and occlusion threshold, respectively.

\textbf{Box-Grained Trajectory Association.}
For associating tracklets between two connected zones with the distance matrix $D$, we propose a novel and effective matching strategy to find all the convincing pairs. Inspired by~\cite{zhong2017re}, all tracklets are associated with the principle of k-reciprocal nearest neighbors. First, we define $N(B_i^h,k)$ as the $k$ nearest neighbors of a probe box $B_i^h$, 
\begin{equation}
\label{equ:assTopk}
    N(B_i^h,k) = (\bar{B}_1, \bar{B}_2,...,\bar{B}_k), |N(B_i^h,k)| = k
\end{equation}
where $|\cdot|$ is the number of top-k candidates. According to \cite{zhong2017re}, the k-reciprocal nearest neighbors are more related to probe box $B_i^h$ than the nearest neighbor. We get the k-reciprocal nearest neighbors $\mathcal{R}(B_i^h, k)$ of $B_i^h$ as:
\begin{small}
\begin{align}
\label{equ:k-reciprocal}
\mathcal{R}(B_i^h, k) = \{\bar{B}_j | (\bar{B}_j \in N(B_i^h, k)) \land (B_i^h \in N(\bar{B}_j, k))\}
\end{align}
\end{small}
Then, for each probe box $B_i^h$ in the exiting zone, we find the matched tracklet by counting the most frequent k-reciprocal nearest frames of a tracklet among all tracklet boxes from the entering zone. Once every box in the exiting zone has been assigned a matched tracklet from the entering zone, two tracklets are assigned the same global ID if the entering tracklet has the most matched boxes that belong to the exiting tracklet.

\begin{algorithm} 
	\caption{Pseudo-code of ICA} 
	\label{alg:ICA} 
	\begin{algorithmic}[1]
	    \renewcommand{\algorithmicrequire}{\textbf{Input:}}
        \renewcommand{\algorithmicensure}{\textbf{Output:}}
		\REQUIRE The trajectories of SCMT in the out cameras $\Tau_{out}$, and in cameras $\Tau_{in}$ , the timestamp threshold for out cameras $T_{out}$, and in cameras $T_{in}$, the zones of out cameras $z_{out}$, and in cameras $z_{in}$ 
		\ENSURE Global Trajectories $\Tau_G$
		\STATE $\Tau_G \gets \varnothing$ 
		\STATE \textit{/$\ast$ Step 1: Trajectories Pool Construction $\ast$/}
		\STATE $\Tau_{out} \gets [\Tau_{out}(t_{out} < T_{out})\cap \Tau_{out}(z_{out})]$
        \STATE $\Tau_{in} \gets [\Tau_{in}(t_{in} > T_{in})\cap  \Tau_{in}(z_{in})] $
		\STATE \textit{/$\ast$ Step 2: Box-Grained Distance Matrix Construction $\ast$/}
		\STATE Construct distance matrix $D$ using equation \ref{equ:disMat} 
		\STATE \textit{/$\ast$ Step 3: Box-Grained Distance Matrix Optimization $\ast$/}
		\STATE $\hat{D} \gets$ Optimize $D$ using the re-ranking algorithm
		\IF{$t_{i,j} < t_{low}$}
		\STATE $\hat{D}_{i,j} \gets e^{\frac{\alpha_t\times(t_{low} - t_{i,j})}{\beta_t}}\times{\hat{D}_{i,j}}$
		\ELSIF{$t_{i,j} > t_{upp}$}
		\STATE $\hat{D}_{i,j} \gets e^{\frac{\alpha_t\times(t_{i,j} - t_{upp})}{\beta_t}}\times{\hat{D}_{i,j}}$
		\ENDIF
		\IF{$r_o > r_{thre}$}
		\STATE $\hat{D}_{i, j} \gets e^{\alpha_o\times(1 + r_o)}\times{\hat{D}_{i,j}}$
		\STATE \textit{/$\ast$ Step 4: Box-Grained Trajectoiries Matching $\ast$/}
		\STATE $\Tau_G \gets$ k-reciprocal nearest neighbors
		\ENDIF
		\STATE $\Tau_G \gets$ post-processing
		\RETURN $\Tau_G$
	\end{algorithmic} 
\end{algorithm}

\section{Experiments}
\label{sec:exp}

\subsection{Datasets and Evaluation Metrics}

\textbf{Datasets.} The CityFlowV2 dataset\cite{tang2019cityflow} is collected from 46 cameras spanning 16 intersections in a mid-sized U.S. city. It covers a diverse set of location types, including intersections, stretches of roadways, and highways. The dataset is divided into six scenarios: three for training, two for validation, and one for testing. The annotations included with the dataset only comprise vehicles that are tracked across multiple cameras. To evaluate the performance of the SCMT module, we manually annotated the complete single-camera tracking results for two cameras and created an additional SCMT dataset. 

To improve the quality of ReID features, our approach uses a combination of real data from the CityFlowV2 dataset, as well as synthetic data generated by VehicleX, a publicly available 3D engine, as reported in \cite{tang2019pamtri}. The training set comprises of 2028 vehicles in total, 666 of which are real vehicles and 1362 are synthetic. And 229345 images are used, 27195 of which are real images and 192150 are synthetic images.

\textbf{Evaluation Metrics and Evaluation Approaches.} For the MCMT, the CityFlowV2 benchmark uses IDF1 score~\cite{ristani2016performance} as the ranking metric. IDF1 measures the ratio of correctly identified detections to the average number of ground-truth and computed detections. For the SCMT, CLEAR metrics~\cite{bernardin2008evaluating} (i.e., Multiple Object Tracking Accuracy (MOTA), ID Switching (IDSW)) and Identity metrics (i.e., IDF1)~\cite{ristani2016performance} are widely accepted to measure the tracking performance. In general, MOTA focuses more on detection performance while IDF1 focuses more on association performance. The HOTA metric~\cite{luiten2021hota} was proposed to provide a better trade-off between detection and association performance. In our experiment, we select IDF1, HOTA, MOTA, and IDSW to evaluate the tracking results from various perspectives. The MCMT metrics are evaluated with the evaluation system of the AICity Challenge 2022\footnote{https://www.aicitychallenge.org/2022-data-and-evaluation/}, whereas the SCMT metrics are evaluated with the TrackEval script \cite{luiten2021hota}.

\subsection{Implementation Details}

\textbf{Detection and ReID Feature Extraction.} The vehicle detection module in our system is based on the Cascade-RCNN algorithm with a SwinTransformer-Base backbone \cite{cai2018cascade,liu2021swin}. This detector is able to detect all vehicles on the scene and provide their locations. The vehicle ReID feature extraction module is an ensemble of five different ReID models, including ResNet-50, ResNeXt101, Res2Net, ConvNext, and HRNet\cite{he2016deep,xie2017aggregated,gao2019res2net,liu2022convnet,wang2020deep}. These models are trained using both cross-entropy loss and triplet loss, and their extracted features are concatenated to form the final ensembled feature representation.

\textbf{Single-Camera Multi-Target Tracking.} For the baseline tracker, detections with a confidence score greater than 0.6 are first matched with previous tracklets. Then, detections with a confidence score between 0.1 and 0.6 are matched with any tracklets that were not matched in the first step. If the cosine distance between the feature vector of the detection box and the feature vector of the tracklet is less than 0.45, the matching is rejected. The lost tracklets are retained for 30 frames in case they reappear. For the Trajectory Re-Link step, the maximum cosine distance threshold is set to 0.4 during the greedy matching process.

\textbf{Inter-Camera Association.} In the tracklet candidates construction step, the zones are manually drawn, as shown in Figure \ref{fig:zones}. For the exiting threshold $T_{out}$ and entering threshold $T_{in}$, different cameras have different threshold values depending on the road structure. In the box-grained distance matrix construction step, we use an occlusion rate threshold of $r_{thre}$ = 0.6 and $\alpha_o$ = 1.1 in Equation \ref{equ:occRefine}. In the box-grained trajectory association step, we use the k-reciprocal nearest neighbors algorithm to match tracklets, with k = 7.

\subsection{Main Results} 
\begin{figure*}[t]
  \centering
\includegraphics[width=1.0\linewidth]{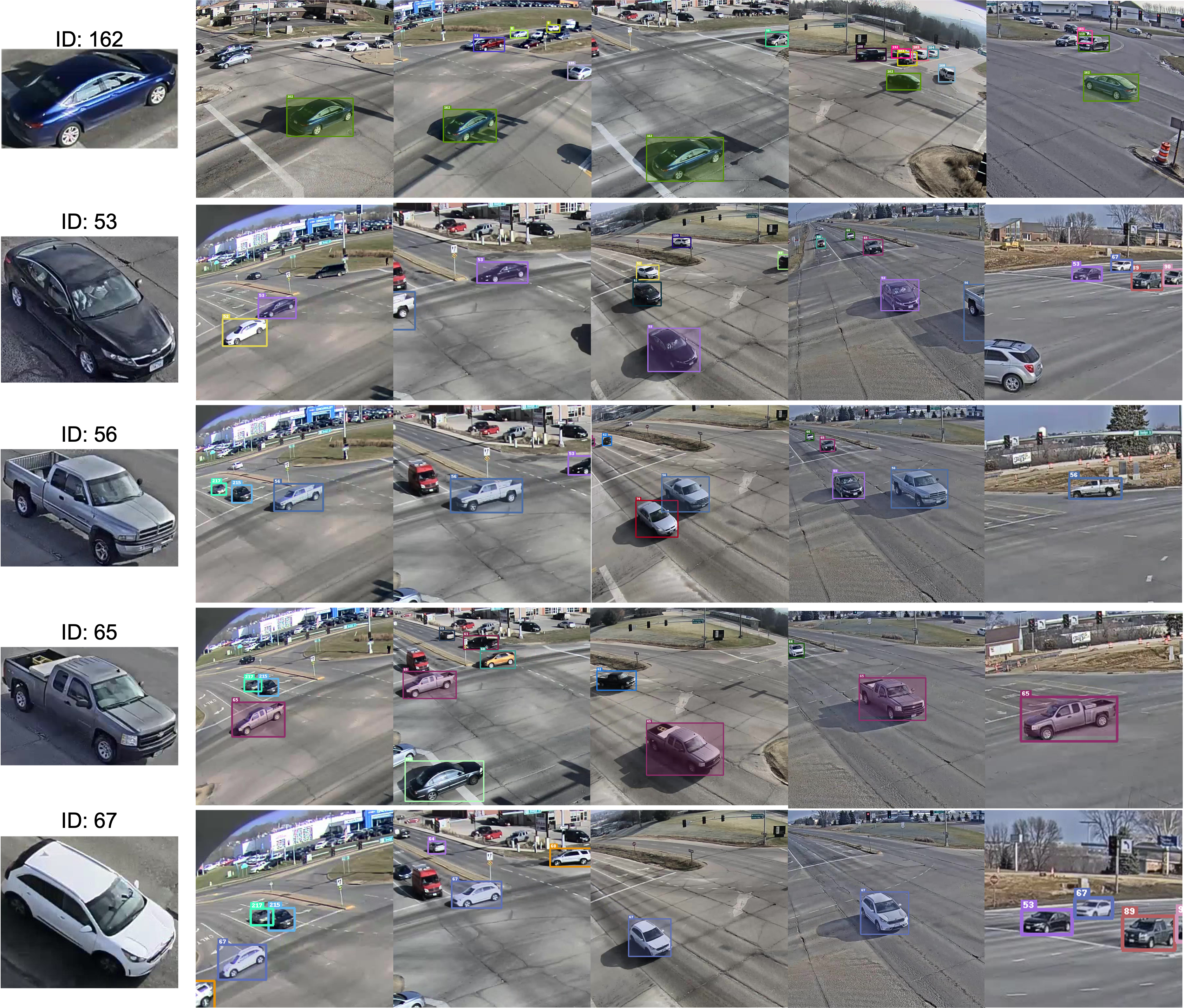}

  \caption{Qualitative results of our MCMT results on CityFlowV2 dataset. Our method is able to accurately maintain the same ID for vehicles as they travel across multiple cameras, even in cases where the vehicle's appearance changes due to lighting or camera angle.}
  \label{fig:visualization}
\end{figure*}

The results depicted in Figure~\ref{fig:visualization} provide a visual representation of how our method performs in terms of Multi-Camera Multi-Target Tracking. The figures demonstrate that our method is able to accurately maintain the same ID for vehicles as they travel across multiple cameras, even in cases where the vehicle's appearance changes due to lighting or camera angle. Our method's performance was evaluated on the CityFlowV2 dataset using the benchmark's evaluation system, and the results were submitted to the 6th AICity Challenge. Our method achieved an IDF1 score of $85.45\%$ and ranked first among all the methods submitted to the benchmark.

\begin{table}[htbp]
    \caption{Leaderboard of CityFlowV2 Benchmark.}
    \centering
	\setlength{\tabcolsep}{2mm}
	\begin{tabular}{c|c|c}  
		\toprule 
		Rank & Team Name & IDF1$\uparrow$ \\  
		\midrule
		1 & CityTrack(ours) & \textbf{85.45} \\
		2 & BOE\cite{ref_boe} & 84.37 \\
		3 & TAG\cite{ref_tag} & 83.71 \\
		4 & FraunhoferIOSB\cite{ref_fraunhofer} & 83.48 \\
		5 & appolo & 82.51 \\
	    6 & Li-Chen-Yi & 82.18 \\
	    7 & Terminus-AI & 81.71 \\
	    8 & FourBeauties & 81.66 \\
	    9 & Orange Peel & 81.40 \\
	    10 & SKKU Automation Lab\cite{ref_skku} & 81.29 \\
		\bottomrule 
	\end{tabular}
	\label{tab:leaderboard} 
\end{table}

Additionally, we compare our Location-Aware SCMT with other methods submitted to the challenge and state-of-art trackers on the SCMT dataset. To ensure fair comparisons, we use the same detector and ReID model for all trackers and supplement the results of the challenge participating teams with their private detections and features. Our method achieves the best performance in all metrics. The results are shown in \ref{tab:scmtOut}.

\begin{table}[htbp]
    \caption{Comparison of the state-of-the-art SCMT methods on the SCMT Dataset. For a fair comparison, methods in the bottom block use detections generated by Cascade RCNN\cite{cai2018cascade} with SwinTransformer-Base Backbone \cite{liu2021swin}. (best in bold).}
    \centering
    \small
	\setlength{\tabcolsep}{2mm}
	\begin{tabular}{lcccc}  
		\toprule 
		Method & IDF1$\uparrow$ & HOTA$\uparrow$ & MOTA$\uparrow$ & IDSW$\downarrow$ \\ 
		\midrule
            \multicolumn{5}{l}{Using Private Detections} \\
            BOE\cite{ref_boe} & 77.97 & 67.29 & 68.75 & 72 \\
            TAG\cite{ref_tag} & 81.94 & 70.08 & 76.52 & 211 \\
            \midrule
            \multicolumn{5}{l}{\makecell[l]{Using Detections Generated by Cascade RCNN\cite{cai2018cascade} with Swin-B \\ Backbone \cite{liu2021swin}}} \\
		DeepSORT\cite{ref_deepsort} & 82.37 & 69.52 & 76.52 & 153 \\
		OCSORT\cite{ref_ocsort} & 85.19 & 73.47 & 77.09 & 70 \\
		ByteTRACK\cite{zhang2021bytetrack} & 86.87 & 72.78 & 80.51 & 68 \\
		StrongSORT\cite{ref_ocsort} & 84.96 & 73.93 & 78.82 & 108 \\
		BOE\cite{ref_boe} & 82.50 & 71.73 & 75.15 & 97 \\
		TAG\cite{ref_tag} & 84.86 & 74.21 & 80.79 & 210 \\
		Location-Aware SCMT & \textbf{89.48} & \textbf{76.75} & \textbf{83.81} & \textbf{33} \\
		\bottomrule 
	\end{tabular}
	\label{tab:scmtOut} 
\end{table}

In summary, our method shows superior performance compared to other methods submitted to the AICity challenge, as well as state-of-the-art trackers, on both single-camera and multi-camera vehicle tracking tasks. The results demonstrate the effectiveness of our CityTrack approach in handling complex traffic scenarios.

\subsection{Ablation Study}

\begin{table}[htbp]
    \caption{Ablation study on the SCMT Dataset for Location-Aware SCMT strategies. i.e., Stationary Sensitive Association (SSA), Trajectory Re-Link (TRL), and Bidirectional Tracking (BT). All results are obtained with the same parameters set. (best in bold).} 
    \centering
    \small
	\setlength{\tabcolsep}{2mm}
	\begin{tabular}{lcccc}
		\toprule 
		Method & IDF1$\uparrow$ & HOTA$\uparrow$ &MOTA$\uparrow$ & IDSW$\downarrow$ \\  
		\midrule
		 baseline & 86.87 & 75.94 & 81.88 & 83 \\
		 baseline+SSA & 89.09 & 76.55 & 82.67 & 56\\
		 baseline+TRL & 87.42 & 76.02 & 81.52 & 68 \\
		 baseline+BT & 87.39 & 76.11 & 82.69 & 82 \\
		 baseline+SSA+TRL & 89.17 & 76.37 & 82.15 & 39 \\
		 \makecell{baseline+SSA+TRL+BT\\(Location-Aware SCMT)} & \textbf{89.48} & \textbf{76.75} & \textbf{83.81} & \textbf{33} \\
		\bottomrule 
	\end{tabular}
	\label{tab:scmt} 
\end{table}

\textbf{Single Camera Multi-target Tracking}
Our ablation study on SCMT aims to evaluate the performance of our proposed strategies and quantify the contribution of each component. We use the same detector, ReID feature extractor, and tracking parameters for all experiments. The results of this ablation study are shown in Table~\ref{tab:scmt}, where we report the contribution of each proposed strategy in our SCMT module. 
\begin{enumerate}[{1)}]
\item SSA: Stationary Sensitive Association leads to a significant improvement for IDF1, indicating that the special consideration for still objects is helpful to maintain the correct identities over time in crowded traffic scenes. 
\item TRL:  Trajectory Re-Link results in an improvement in IDF1 and IDSW, implying that it makes the broken trajectories more complete. 
\item BT: Bidirectional Tracking improves MOTA by boosting the bounding boxes recall. 
\item SSA+TRL: The IDF1 and IDSW further improved after using the Stationary Sensitive Association module and the Trajectory Re-Link module. 
\item SSA+TRL+BT: It achieves the best tracking results by combining the three strategies.
\end{enumerate}

\begin{table}[htbp]
    \caption{Comparison of different distance matrix construction and trajectory association methods. Where ``Hun'' and ``K-Rec'' represent the Hungarian method and k-reciprocal nearest neighbors respectively. (best in bold).}
    \centering
    \small
	\setlength{\tabcolsep}{2mm}
	\begin{tabular}{c|cc|ccc}  
		\toprule 
		Method & Hun & K-Rec & IDF1$\uparrow$ & IDP$\uparrow$ & IDR$\uparrow$ \\  
		\midrule
		tracklet-wise & \checkmark & & 82.39 & 84.97 & 79.96 \\
		tracklet-wise & & \checkmark & 84.16 & \textbf{92.70} & 77.07 \\
		box-grained & \checkmark & & 80.16 & 85.42 & 75.50 \\
		box-grained & & \checkmark & \textbf{85.45} & 91.15 & \textbf{80.42} \\
		\bottomrule 
	\end{tabular}
	\label{tab:ass} 
\end{table}

\begin{table}[htbp]
    \caption{Comparison of box-Grained distance matrix optimization methods. Where ``RR'', ``TTW'', and ``OR'' represent the re-ranking, traveling time window, and occlusion rate respectively. (best in bold).}
    \centering
    \small
	\setlength{\tabcolsep}{2mm}
	\begin{tabular}{ccc|ccc}  
		\toprule 
		RR & TTW & OR & IDF1$\uparrow$ & IDP$\uparrow$ & IDR$\uparrow$ \\  
		\midrule
		& & & 84.49 & 90.97 & 78.87 \\
		\checkmark & & & 85.23 & 91.21 & 79.99  \\
		\checkmark & \checkmark & & 85.39 & 91.15 &	80.31  \\
		\checkmark & & \checkmark & 85.29 & \textbf{91.27} & 80.05  \\
		\checkmark & \checkmark & \checkmark & \textbf{85.45} & 91.15 & \textbf{80.42}  \\
		\bottomrule 
	\end{tabular}
	\label{tab:optim} 
\end{table}

\begin{table}[htbp]
    \caption{Comparison of different ``k'' values in k-reciprocal nearest neighbors association. (best in bold).}
    \centering
    \small
	\setlength{\tabcolsep}{2mm}
	\begin{tabular}{c|ccc}
		\toprule 
		k & IDF1$\uparrow$ & IDP$\uparrow$ & IDR$\uparrow$ \\  
		\midrule
		3 & 84.45 & 90.99 & 78.78 \\
		5 & 85.15 & \textbf{91.60} & 79.56 \\
		7 & \textbf{85.45} & 91.15 & \textbf{80.42} \\
		9 & 84.85 & 90.74 & 79.68 \\
		\bottomrule 
	\end{tabular}
	\label{tab:topk} 
\end{table}

\textbf{Inter-Camera Association.} We perform ablation studies on the ICA module to investigate the effect of the proposed Distance Matrix Construction and Trajectory Association method, as well as the effect of distance matrix optimization strategies and the ``k'' value of the k-reciprocal nearest neighbors association.

Table~\ref{tab:ass} presents a comparison of different distance matrix construction and trajectory association methods. We used the Hungarian and k-reciprocal nearest neighbors algorithms for both tracklet-wise and box-grained matrix construction methods, and the results indicate that the combination of box-grained distance matrix with the k-reciprocal nearest neighbors association algorithm is the best for achieving an IDF1 score of 85.45\% on the CityFlowV2 dataset.

Table~\ref{tab:optim} shows a comparison of optimization methods applied to the box-grained distance matrix. From the results we can see that re-ranking, using a traveling time window, and including an occlusion rate in the optimization of the distance matrix all improve the performance of the overall MCMT results.

In our Box-Grained Matching method, ``k'' in the k-reciprocal nearest neighbors association is a critical hyperparameter. Here, we perform ablation studies to investigate how ``k'' values affect the tracking performance. As shown in Table~\ref{tab:topk}, the choice of k = 7 gives the highest IDF1 score. Note that, although the variation of k affects the MCMT performance remarkably, using the Hungarian matching can only achieve 80.16\% of IDF1 for the box-grained method, which is lower than using any of the above k values.

\section{Conclusion}
\label{sec:con}

In this paper, we proposed a novel Multi-Camera Multi-Target Tracking framework that is designed to handle city-scale MCMT tasks. The proposed method consists of several key components including vehicle detection, ReID feature extraction, Single-Camera Multi-Target Tracking, and Inter-Camera Association, which together enable MCMT results. We proposed two key innovations to improve the tracking performance: our Location-Aware Tracking method, which considers the spatial context of targets, and our Box-Grained Matching method, which performs a fine-grained comparison of bounding-box-level features to improve Inter-Camera Association accuracy.

We evaluated the performance of our method on the public test set of the CityFlowV2 dataset. Our method achieved an IDF1 score of 85.45\%, which is the highest score achieved on the benchmark, setting a new record for the task of city-scale multi-camera multi-target tracking.

{\small

\bibliographystyle{ieee_fullname}
\bibliography{bare_jrnl_new_sample5}
}

\vfill

\end{document}